\documentclass{article}

\PassOptionsToPackage{numbers, compress}{natbib}
\usepackage[preprint]{neurips_2026}

\usepackage[utf8]{inputenc} 
\usepackage[T1]{fontenc}    
\usepackage{hyperref}       
\usepackage{url}            
\usepackage{booktabs}       
\usepackage{amsfonts}       
\usepackage{nicefrac}       
\usepackage{microtype}      
\usepackage{xcolor}         

\usepackage{graphicx}
\usepackage{algorithm}
\usepackage{algpseudocode}
\usepackage{amsmath}

\usepackage{adjustbox, makecell}
\usepackage{multirow}
\usepackage{subcaption}

\title{Sparse Code Uplifting for Efficient 3D Language Gaussian Splatting}

\author{%
  \textbf{Lovre Antonio Budimir}\textsuperscript{1,2,}\thanks{Corresponding author: lovre-antonio.budimir@fer.unizg.hr}
  \quad
  \textbf{Yushi Guan}\textsuperscript{2,3}
  \quad
  \textbf{Steve Ryhner}\textsuperscript{2,3}
  \\
  \textbf{Sven Lon\v{c}ari\'{c}\textsuperscript{1}}
  \quad
  \textbf{Nandita Vijaykumar\textsuperscript{2,3}}
  \\[0.8em]
  \textsuperscript{1}Faculty of Electrical Engineering and Computing,
  University of Zagreb
  \\
  \textsuperscript{2}Department of Computer Science,
  University of Toronto
  \\
  \textsuperscript{3}Vector Institute
}

\begin{document}

\maketitle

\begin{abstract}
    3D Language Gaussian Splatting (3DLGS) augments 3D Gaussian Splatting with language-aligned visual features for open-vocabulary 3D scene understanding. A core challenge is efficiently associating high-dimensional vision-language embeddings with millions of 3D Gaussians while preserving efficient feature rendering for text-based querying. Existing methods either store dense features directly on Gaussians, causing high storage costs and slow rendering, or learn compact representations through expensive per-scene optimization with repeated feature rasterization. No existing method simultaneously achieves fast 3D semantic reconstruction, efficient storage, and fast rendering. We propose \textbf{SCOUP} (\textbf{S}parse \textbf{CO}de \textbf{UP}lifting), which addresses all three by decoupling language representation learning from 3D Gaussian optimization. Rather than working directly in 3D, we learn sparse codebook-based representations entirely using features associated with 2D image regions, associating each region with a sparse set of codebook coefficients. We then uplift these coefficients to 3D Gaussians with our weighted sparse aggregation using Gaussian-to-pixel associations, where each Gaussian accumulates coefficients over codebook atoms across views. Top-$K$ filtering then extracts the most dominant multi-view coefficients per Gaussian, enabling efficient storage and fast rendering. Our method achieves up to 400× training speedup while being 3× more memory efficient during training compared to the state-of-the-art in rendering speed. Across multiple benchmarks, SCOUP matches or outperforms existing methods in open-vocabulary querying accuracy.
  
\end{abstract}
\section{Introduction}
\label{sec:intro}

3D Gaussian Splatting (3DGS) has emerged as a powerful tool that enables high-quality reconstruction of 3D scenes from multi-view 2D images~\cite{3DGS}. 3DGS represents scene appearance and geometry with an explicit set of anisotropic Gaussians that can be rasterized in real time on modern GPUs. Beyond these visual and spatial properties, these 3D representations are increasingly extended with language-aligned visual features extracted from vision–language models (VLMs) such as CLIP~\cite{clip,openclip}. This creates 3D Language Gaussian Splatting (3DLGS) scenes built on top of 3DGS that can be  queried with arbitrary text prompts. Recent methods~\cite{langsplat,LEGaussians,langsplatv2} demonstrate that 3DLGS enables open-vocabulary 3D understanding and language-guided interaction, with valuable applications in language-driven navigation, manipulation, and interaction in robotics and AR/VR settings~\cite{VLMaps,ConceptFusion,D3Fields,vrsplat}.

3DLGS aims to build a 3D representation that, when rendered, produces high-dimensional features comparable with text embeddings to enable open-vocabulary scene querying. Typical pipelines use SAM~\cite{sam} to segment each scene image into regions, then CLIP~\cite{clip} to generate \(512\)-dimensional embeddings per region~\cite{LEGaussians,langsplat}. These features are then optionally compressed (e.g., via a scene-specific autoencoder) before being used for training or directly uplifting to 3D Gaussians. The Gaussians' location, covariance, and opacity are inherited from the pre-trained 3DGS~\cite{langsplat,ludvig,cf3}. During rendering, the features stored in each Gaussian are rasterized, and if compression was applied, an additional decoding step is required to recover the original feature for text-based feature comparison and querying~\cite{langsplat,langsplatv2}.

There are two major challenges in 3DLGS scene reconstruction and rendering. First, 3D semantic reconstruction can incur significant latency and/or memory usage depending on the technique used. One commonly adopted approach is to directly uplift high-dimensional features from 2D images to 3D Gaussians~\cite{ludvig,occams,vala}, leveraging Gaussian-to-pixel associations readily available from the geometric and visibility priors of a pretrained 3DGS model. Despite being fast, this process can lead to high memory usage when Gaussian-pixel associations are complex, and the resulting 3DLGS model incurs high storage costs due to millions of Gaussians each storing a full high-dimensional feature vector. Alternatively, training-based methods that compress high-dimensional features using scene-specific autoencoders~\cite{langsplat} successfully mitigate memory overhead, but incur significantly longer 3D reconstruction runtimes due to the computational burden of repeated feature rasterization during optimization. 
Second, rendering is typically slow due to either the need to rasterize high-dimensional features directly, or the need to decode rasterized low-dimensional features with a neural network~\cite{occams,feature3dgs,langsplat}. LangSplatV2~\cite{langsplatv2} addresses this by associating each Gaussian with sparse coefficients over a shared high-dimensional codebook, replacing the expensive neural decoder with a single matrix multiplication between the coefficients and the codebook. LangSplatV2 uses the high-dimensional embeddings from 2D images as ground-truth supervision, jointly optimizing the codebook and sparse Gaussian coefficients. However, this results in slow training due to the difficulty of jointly optimizing the codebook and sparse coefficients alongside repeated rasterization during training. One alternative would be to first uplift high-dimensional features from 2D to 3D, then train the codebook and sparse coefficients against these uplifted 3D features. However, our investigations found this degrades accuracy as Gaussians associated with larger objects outnumber those associated with smaller ones. This causes larger objects to dominate the codebook updates,  resulting in poor feature representation for small objects. In summary, \emph{existing works do not achieve fast 3D semantic reconstruction, efficient storage, and fast rendering simultaneously}.

\begin{figure}[t]
  \centering
  \includegraphics[width=\linewidth]{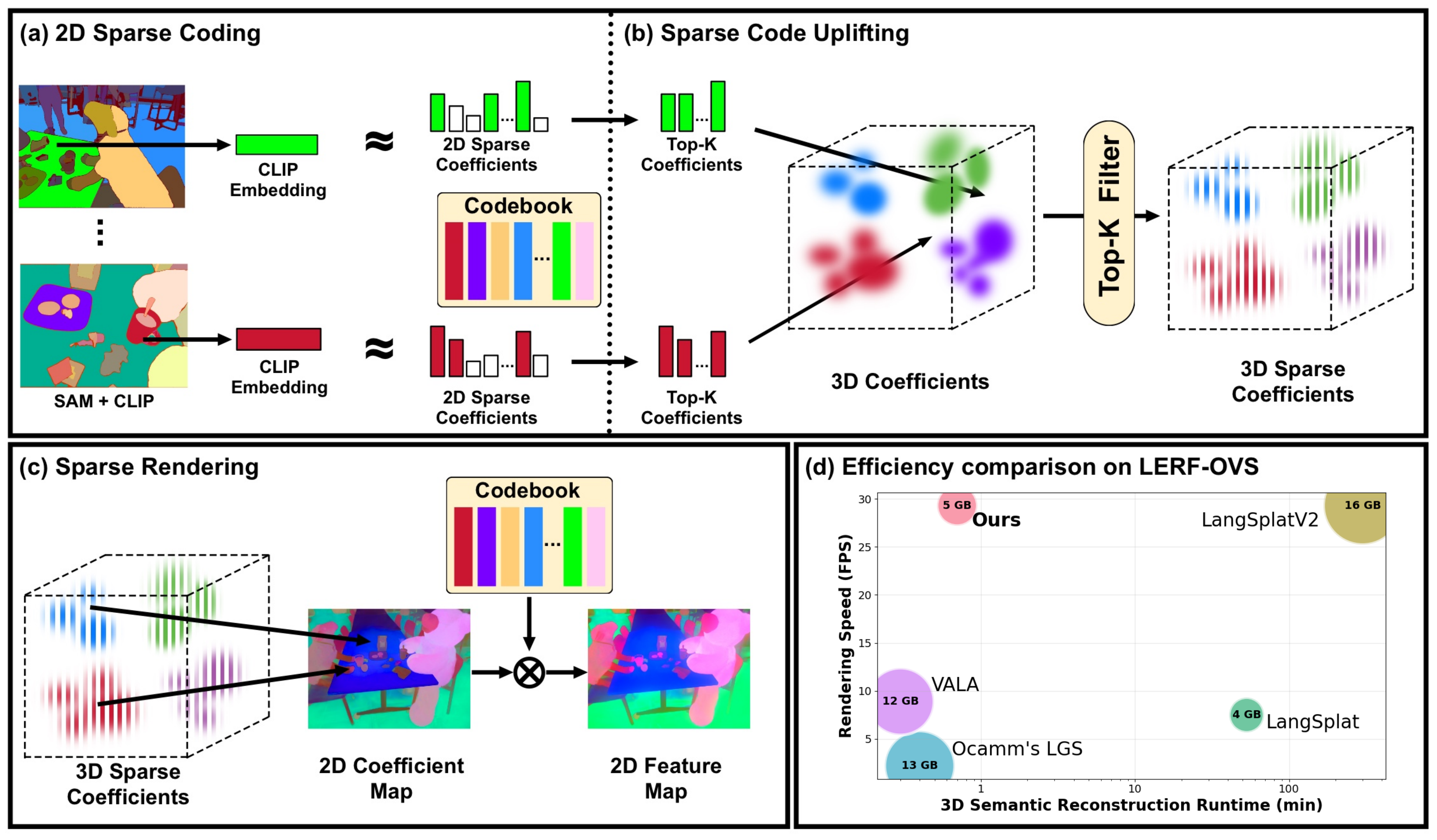}
  \caption{\textbf{SCOUP} reconstructs 3D semantic scene representations in three stages: (a) jointly optimizing a scene-aware codebook and sparse coefficient maps to represent dense 2D CLIP features; (b) efficiently uplifting only non-zero coefficients to 3D Gaussians and applying top-$K$ filtering to enforce multi-view consistency by retaining only the most dominant coefficients across views. During inference (c), the sparse codes stored in each Gaussian enable fast language feature rendering, ensuring the best overall efficiency (d) for 3D Language Gaussian Splatting.}
  \label{fig:pipeline}
\end{figure}

Our goal in this work is to simultaneously enable fast 3D semantic reconstruction, efficient storage, and fast rendering for 3DLGS, while achieving state-of-the-art accuracy. We propose \textbf{S}parse \textbf{Co}de \textbf{Up}lifting (SCOUP) for efficient 3DLGS. Rather than optimizing language representations directly in 3D Gaussian space, our key idea is to first jointly train a shared high-dimensional codebook and sparse coefficients in CLIP's native 2D space, where each SAM-segmented region is associated with a sparse set of codebook coefficients to reconstruct its CLIP embedding (Fig.~\ref{fig:pipeline}(a)). We then uplift the learned 2D region coefficients to 3D Gaussians by our weighted sparse aggregation: by assigning all pixels within a region the same set of sparse coefficients, the Gaussian-to-pixel associations from a pretrained 3DGS model can be efficiently used to aggregate coefficients from 2D to 3D without any additional optimization. As each Gaussian aggregates contributions across multiple views, the accumulated coefficients implicitly form a semantic voting system: high-weight coefficients represent confident consensus, while low-weight coefficients represent multi-view noise. To denoise the uplifted coefficients, we retain only the top-\(K\) coefficients by total aggregated weight, filtering the rest to zero (Fig.~\ref{fig:pipeline}(b)). This ensures each Gaussian stores only \(K\) coefficient weights, enabling efficient storage and fast rendering (Fig.~\ref{fig:pipeline}(c)). 

Our method achieves fast 3D semantic reconstruction by exploiting three properties. First, SAM~\cite{sam} segments each scene into only a limited number of regions, keeping the 2D coefficient optimization significantly smaller than optimizing coefficients directly in 3D across millions of Gaussians. Second, we avoid the expensive repeated rasterization required by existing works \cite{langsplatv2,feature3dgs,langsplat}, which arises from optimizing coefficients associated with 3D Gaussians against 2D image embedding supervision. Third, the learned sparse coefficients are efficiently uplifted to 3D Gaussians requiring no additional optimization and no high-dimensional feature rasterization. Our 3D semantic reconstruction is also highly memory efficient: each Gaussian aggregates and stores only sparse coefficients instead of full high-dimensional feature vectors, avoiding the costly high-dimensional aggregation or optimization across millions of Gaussians that prior works require. Our method also maintains or outperforms existing works in accuracy, as the codebook is trained to directly reconstruct 2D CLIP features, which is the representation that text embeddings are compared against at query time. Our sparse coefficient uplifting is nearly lossless given accurate Gaussian-to-pixel associations, and as we will demonstrate experimentally, treating multi-view aggregation as a voting system allows our top-\(K\) selection to act as an effective noise filter. By discarding low-weight coefficients, we intrinsically improve multi-view consistency and open-vocabulary querying performance.

We demonstrate the effectiveness of our method across three datasets: LERF-OVS~\cite{LERF2023}, 3D-OVS~\cite{3dovs2023}, and Mip-NeRF360~\cite{mipnerf360}. Compared to LangSplatV2, our method achieves more than \(400\times\) 3D semantic reconstruction speedup while maintaining identical storage and rendering speed, and is \(3\times\) more efficient in training memory. Compared to the fastest uplifting-based methods~\cite{vala,occams}, our reconstruction time remains under a minute, while rendering at least \(3\times\) faster (Fig.~\ref{fig:pipeline}(d)).
In terms of accuracy, we outperform LangSplatV2~\cite{langsplatv2} and VALA~\cite{vala}, the leading optimization-based and uplifting-based 3DLGS methods by +4.5\% and +2.7\% in mIoU, respectively, on open-vocabulary segmentation on LERF-OVS.

In summary, our key contributions are:
\textbf{(1)} To our knowledge, we propose the first framework that encodes 2D high-dimensional dense features into 2D sparse coefficient maps and uplifts them to a 3DGS scene for efficient 2D-to-3D and 3D-to-2D transfer and storage of language features. 
\textbf{(2)}~We propose two novel techniques: (i) 2D Sparse Coding that jointly trains a shared codebook and per-region sparse codes to reconstruct CLIP features and (ii) Sparse Code Uplifting formulated as a weighted voting process over 2D codebook atoms guided by 3DGS priors, followed by top-K consensus filtering for compact storage, efficient rendering and semantic noise reduction.
\textbf{(3)} Our method achieves more than \(400\times\) training speedup and uses a third of the training memory compared to LangSplatV2, while maintaining its rendering speed and storage, and matches or outperforms other baselines on 3D open-vocabulary querying across benchmarks.

\section{Related Work}
\label{sec:literature}

\paragraph{3D Gaussian Splatting.} In contrast to the implicit representations used in Neural Radiance Fields (NeRFs) \cite{nerf}, 3D Gaussian Splatting (3DGS) \cite{3DGS} has emerged as a high-efficiency alternative for 3D scene reconstruction from 2D images. By representing 3D space through explicit anisotropic Gaussian primitives and leveraging a differentiable rasterization pipeline, 3DGS achieves real-time rendering speeds. This has motivated the research community to extend 3DGS capabilities to support various open-vocabulary tasks in 3D space, from editing \cite{gaussianeditor,gaussianeditor_del,dreamcatalyst,locgaussianediting,3dsceneeditor} and generating \cite{gsgen,dreamgaussian,luciddreamer} objects using textual prompts, to embedding language representations into 3DGS to support open-vocabulary querying \cite{langsplat,feature3dgs,LEGaussians}. 

\paragraph{3D Language Gaussian Splatting.} 
Transferring knowledge from 2D vision-language models (VLMs) \cite{clip,lseg,dinotxt} into 3D scene representations is a critical step for unlocking open-vocabulary capabilities. Due to the explicit nature of 3DGS, storing high-dimensional features within individual Gaussians has led to two primary paradigms for 2D-to-3D feature transfer: optimization-based \cite{langsplat,feature3dgs,LEGaussians,df-3dgs,langsurf,langsplatv2,goi,ccl-lgs} and optimization-free \cite{ludvig,drsplat,occams,vala} methods.

Optimization-based methods learn language features through a differentiable rasterization pipeline~\cite{3DGS}. To speed up training and reduce the memory footprint, LangSplat~\cite{langsplat} trains a scene-specific autoencoder to compress CLIP embeddings into a low-dimensional latent space, which is then used to distill knowledge into 3D Gaussians. While this extreme compression reduces memory requirements, decoding the rendered latent representations back to the original CLIP space results in slow inference and a loss of semantic information. LangSplatV2 \cite{langsplatv2} and LEGaussians~\cite{LEGaussians} adopt codebook-based approaches where Gaussians reference a global feature dictionary. Specifically, LEGaussians quantizes 2D features into a discrete codebook, then learns compact per-Gaussian features that can be decoded to codebook-index predictions. In contrast, LangSplatV2 \cite{langsplatv2} represents each Gaussian as a weighted combination of multiple codebook entries. While this offers greater representational capacity than \cite{LEGaussians}, LangSplatV2 suffers from the computationally costly joint optimization of codebook entries and Gaussian coefficients through the rasterization pipeline. While codebook-based formulations have been explored in \cite{goi,ccl-lgs,gala,langsplatv2,LEGaussians}, no prior work leverages the codebook to construct 2D sparse coefficient maps. 

Recent works propose training-free feature uplifting to pre-trained 3DGS models by leveraging their geometric and visibility priors ~\cite{ludvig,drsplat,occams,vala}. LUDVIG~\cite{ludvig} performs a weighted multi-view aggregation of 2D features with a reported speed of 2ms per image and feature dimension. Dr. Splat~\cite{drsplat} uses a similar formulation but assigns language representations only to the most dominant Gaussians along a viewing ray. To reduce storage requirements and improve scalability, Dr. Splat utilizes Product Quantization (PQ) to store features as compact representations. Both \cite{drsplat} and \cite{ludvig} query objects directly in 3D space prior to rendering the results. VALA~\cite{vala} speeds up feature uplifting by filtering out occluded and low-visibility Gaussians during the aggregation procedure, yet it uplifts and renders full-dimensional CLIP features. Occam’s LGS~\cite{occams} adds support for arbitrary feature dimensionality uplifting with scene-specific autoencoders, as in LangSplat~\cite{langsplat}. On the other hand, $\text{CF}^3$~\cite{cf3} moves the autoencoder directly into 3D space, learning low-dimensional latent representations for every Gaussian, which increases semantic reconstruction time. To further improve the efficiency of 3DGS, methods like \cite{cf3} and \cite{ludvig} rely on filtering or adaptive sparsification to reduce the memory footprint. In contrast, our method optimizes the CUDA implementation of feature uplifting \cite{ludvig} to support sparse codes, enabling efficient uplifting of high-dimensional language representations to a 3DGS scene and supporting LangSplatV2 level rendering throughput \cite{langsplatv2} in a fraction of the time.

\section{Preliminaries}
\label{sec:preliminaries}

\textbf{3D Gaussian Splatting (3DGS)}
represents a scene as a large collection of anisotropic 3D Gaussians~\cite{3DGS}. Each Gaussian is defined by a 3D center coordinate \(\boldsymbol\mu_i \in \mathbb R^{3}\),
a 3D size scaling factor \(\mathbf s_i \in \mathbb R^{3}\), a rotation quaternion \(\mathbf q_i \in \mathbb R^{4}\), a view‑dependent color \(\mathbf c_i(v) \in \mathbb R^{3}\) described via spherical‑harmonic coefficients, and an opacity value \(o_i \in \mathbb R\). All parameters are learnable and are collectively denoted as 
\[
\boldsymbol\Theta = \{\,\boldsymbol\mu_i,\;\mathbf s_i,\;\mathbf q_i,\;\mathbf c_i(\cdot),\;o_i\;\}_{i=1}^N
\]
for a scene with \(N\) Gaussian points.

Optimization of 3DGS parameters relies on a differentiable tile-based rasterizer to project 3D Gaussians onto a 2D image~\cite{3DGS}. Following~\cite{KPLD21,3DGS}, given a camera pose with viewing direction \(v\), the estimated color of pixel \(p\) in an image \(\mathbf I_v \in \mathbb R^{H\times W\times3}\) is defined as:
\begin{equation}
\mathbf I_v(p) = \sum_{i\in\mathcal N_{v,p}} \mathbf c_i(v)\;e_i(v,p),
\label{eq:alpha-blending}
\end{equation}
where \(\mathcal N_{v,p}\) denotes the set of Gaussians whose projected splats overlap pixel \(p\), ordered from front to back along viewing ray \(v\), and \(e_i(v,p)\) is the contribution weight of the \(i\)-th Gaussian in this ordered list. The weights are obtained via standard \(\alpha\)-blending~\cite{alpha_blending}:
\begin{equation}
e_i(v,p) = \alpha_i(v,p)\;\prod_{j=1}^{\,i-1} \bigl(1 - \alpha_j(v,p)\bigr),
\label{eq:alpha-blending-decomposed}
\end{equation}
where \(\alpha_i(v,p) = o_i\,G_i^{2D}(v,p)\) is the alpha value of the \(i\)-th Gaussian given its opacity \(o_i\) and projected 2D footprint \(G_i^{2D}(v,p)\). 

\textbf{Feature Uplifting}
~\cite{ludvig,drsplat} is a training-free direct registration of language features from 2D feature maps to the 3DGS scene. Following~\cite{ludvig}, \(\mathcal{S}_i\) denotes the set of all view-pixel pairs \((v,p)\) to which the \(i\)-th Gaussian contributes during rendering. The language feature vector \(\mathbf{f}_i \in \mathbb{R}^{512}\) for the \(i\)-th Gaussian is defined as:
\begin{equation}
\mathbf{f}_i = \frac{1}{{Z_i}}\sum_{(v,p) \in \mathcal{S}_i} e_i(v,p) \mathbf{F}_{v}(p),
\label{eq:uplift}
\end{equation}
where \(Z_i = \sum_{(v,p) \in \mathcal{S}_i} e_i(v,p)\) is the normalization factor, and \(\mathbf{F}_{v}(p) \in \mathbb{R}^{D}\) is the semantic feature vector at pixel \(p\) in view \(v\). Storing a full \(D\)-dimensional feature vector per Gaussian scales storage linearly with \(D\), and directly replacing the color term \(\mathbf{c}_i(v)\) in \eqref{eq:alpha-blending} with \(\mathbf{f}_i\) similarly increases rendering time proportional to the feature dimension.

\section{Methodology}
\label{sec:method}

\subsection{2D Sparse Coefficient Maps}
\label{sec:sparse_coeffient_maps}

Following prior work~\cite{langsplat, langsplatv2}, we extract high-dimensional 2D semantic feature maps using SAM~\cite{sam} and CLIP~\cite{clip}. Specifically, for each input image \(\mathbf{I}_v\), SAM generates a set of non-overlapping binary masks. Each masked region is encoded with CLIP to obtain a semantic feature vector, which is then assigned to all pixels within the corresponding mask, yielding a pixel-wise semantic feature map \(\mathbf{F}_v \in \mathbb{R}^{H \times W \times D}\). Following existing work, we repeat this process at three SAM granularity levels: fine, medium, and coarse, to capture multi-scale semantic information. For notational clarity, the remainder of this section describes our method with respect to a single semantic level.

The semantic representation of a single 3D scene is highly redundant and occupies only a small subspace of the full CLIP feature space~\cite{LEGaussians}, making sparse coding a natural fit for compact scene-specific feature representation. Unlike autoencoder-based compression~\cite{langsplat,feature3dgs}, which requires a neural decoder at render time, sparse coefficients can be decoded via a simple matrix multiplication with the codebook. We therefore encode dense 2D feature maps into 2D sparse coefficient maps, where each pixel is represented as a sparse convex combination of at most \(K\) basis vectors from a learned scene-aware codebook, restricting each pixel to at most \(K\) non-zero coefficient values.

We formulate this as a 2D sparse coding problem, learning a scene-aware codebook \(\mathbf{C} \in \mathbb{R}^{L \times D}\) of \(L\) basis vectors. Each pixel-wise feature \(\mathbf{F}_v(p) \in \mathbb{R}^{D}\) is approximated as a sparse convex combination of at most \(K\) basis vectors:
\begin{equation}
    \mathbf{F}_v(p) \approx \mathbf{W}_v(p)\; \mathbf{C}, \quad \text{s.t.} \quad \|\mathbf{W}_v(p)\|_0 \leq K, \quad \mathbf{W}_v(p) \geq \mathbf{0}, \quad \mathbf{1}^\top \mathbf{W}_v(p) = 1,
    \label{eq:sparse_coding_vector}
\end{equation}
where \(\mathbf{W}_v(p) \in \mathbb{R}^L\) is the sparse coefficient vector at pixel \(p\). Since CLIP features are constant within a SAM region, we parameterize coefficients per region rather than per pixel, so all pixels in a region share \(\mathbf{W}_v(p)\). To enforce these constraints, we apply a normalized soft top-\(K\) softmax over learnable region-wise logits, optimizing these logits jointly with the codebook via a cosine reconstruction loss.

Crucially, while LangSplatV2~\cite{langsplatv2} couples codebook learning with 3DGS optimization, our sparse coding operates entirely in image space, prior to 3D lifting, decoupling semantic compression from the 3DGS pipeline. This places the scene-aware codebook in the same 2D domain on which CLIP~\cite{clip} is trained. The number of extracted masked regions is much smaller than the number of Gaussians, yielding fewer features to compress, while region-based CLIP supervision ensures that small and large objects contribute with more balance to the codebook optimization. 

\subsection{Sparse Coefficient Uplifting}
\label{sec:sparse_coeffient_uplift}

By exploiting the sparsity of our 2D coefficient maps, we uplift only the \(K\) non-zero coefficient values to 3D Gaussians rather than full \(D\)-dimensional feature vectors. This bounds the computational cost of the lifting process strictly by \(K\), independent of the original feature dimension \(D\), yielding a \(D/K\) speedup over dense feature uplifting (\(K \ll D\)). 

We reformulate the dense feature uplifting from \eqref{eq:uplift} as sparse coefficient uplifting:
\begin{equation}
\mathbf{w}_i = \frac{1}{Z_i}\sum_{(v,p) \in \mathcal{S}_i} e_i(v,p)\; \mathbf{W}_{v}(p)
\label{eq:coeffient_uplift}
\end{equation}
where \(\mathbf{w}_i \in \mathbb{R}^L\) is the aggregated coefficient vector assigned to the \(i\)-th 3D Gaussian. Since each \(\mathbf{W}_v(p)\) contains at most \(K\) non-zero elements, the weighted aggregation operates only over non-zero entries, reducing the time complexity of uplifting from \(\mathcal{O}(|\mathcal{S}_{i}|D)\) in \eqref{eq:uplift} to \(\mathcal{O}(|\mathcal{S}_{i}|K)\) per Gaussian.

As each viewpoint contributes a different set of active coefficient indices, the magnitude of each entry in \(\mathbf{w}_i\) reflects cross-view consensus: codebook atoms supported by many views accumulate large weights, while atoms activated by only a few remain small. We therefore restore strict sparsity by retaining the \(K\) dominant coefficients of \(\mathbf{w}_i\) and setting the rest to zero, treating top-\(K\) filtering as a semantic voting step that consolidates view-consistent signal and suppresses noisy contributions. 

The final sparse coefficient vector \(\hat{\mathbf{w}}_i\) is computed as:
\begin{equation}
    \hat{\mathbf{w}}_i = \frac{\text{Top}_K(\mathbf{w}_i)}{\|\text{Top}_K(\mathbf{w}_i)\|_1}
    \label{eq:topk_norm}
\end{equation}
where \(\text{Top}_K(\cdot)\) retains only the \(K\) dominant elements and the normalization ensures the resulting sparse weights sum to one. Note that this final normalization subsumes the per-Gaussian scalar \(Z_i\), which can therefore be omitted in practice. This filtering and contribution consolidation mechanism allows each 3D Gaussian to store only \(K\) coefficient values and their corresponding indices, resulting in a compact memory footprint, efficient rendering, and better multi-view consistency.

\subsection{Sparse Coefficient Rendering}

By storing only \(K\) non-zero coefficient values per Gaussian, rendering operates on sparse coefficients rather than full \(D\)-dimensional feature vectors, accessing only non-zero entries during weighted aggregation. Sparse coefficient rendering is defined as
\begin{equation}
\hat{\mathbf{W}}_v(p) = \sum_{i\in\mathcal N_{v,p}} \hat{\mathbf{w}}_i \;e_i(v,p),
\label{eq:alpha-blending-sparse}
\end{equation}
where \(\hat{\mathbf{W}}_v(p)\) is the projected sparse coefficient map on the 2D plane for view \(v\). This sparse aggregation is directly supported by the rendering pipeline of \cite{langsplatv2}, which efficiently accesses only non-zero coefficient entries via their corresponding sparse indices.
To recover a 2D semantic feature map of CLIP embeddings from the rendered sparse coefficient map, we perform a simple matrix multiplication with the scene-aware codebook \(\mathbf{C}\) learned in Section \ref{sec:sparse_coeffient_maps}:
\begin{equation}
\mathbf{F}_v(p) \approx \hat{\mathbf{W}}_v(p) \; \mathbf{C},
\label{eq:clip-reconstruction}
\end{equation}
where the reconstructed \(\mathbf{F}_v(p)\) can be directly compared with CLIP text embeddings to support open-vocabulary scene querying.

We perform all steps of our pipeline independently for each of the three SAM semantic levels, and render all three sparse coefficient maps to 2D space simultaneously, following \cite{langsplatv2}.

\section{Experiments}
\label{sec:experiments}

\subsection{Datasets and Metrics}

We evaluate our approach on three datasets for 3D open-vocabulary scene understanding. The \textbf{LERF-OVS}~\cite{LERF2023} dataset consists of four in-the-wild scenes captured using the Polycam application on an iPhone. LangSplat~\cite{langsplat} extends the LERF dataset by adding ground-truth segmentation masks and localization points for textual queries. Following previous works~\cite{langsplat, langsplatv2}, we report localization accuracy for open-vocabulary localization and IoU for segmentation. We further evaluate segmentation on the \textbf{Mip-NeRF360}~\cite{mipnerf360} and \textbf{3D-OVS}~\cite{3dovs2023} datasets. For Mip-NeRF360, we use ground-truth annotations provided by GAGS~\cite{gags} and report IoU on four indoor and outdoor scenes. For 3D-OVS, we report segmentation on five scenes with long-tail objects: bed, bench, room, sofa, and lawn. We also conduct a thorough analysis of semantic reconstruction runtime and memory requirements on all three datasets. 

\begin{figure}[t]
  \centering
  \includegraphics[width=\linewidth]{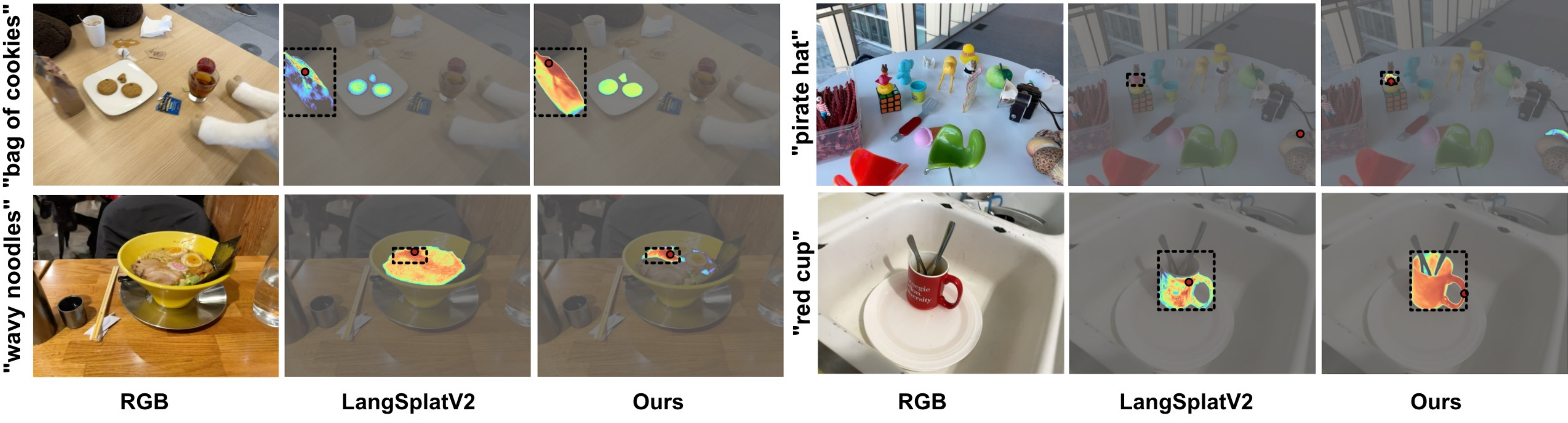}
  \caption{Qualitative results for open-vocabulary 3D object localization on the LERF dataset.}
  \label{fig:qualitative_loc}
\end{figure}

\subsection{Implementation Details}

The input to our model is a 3DGS scene trained for 30,000 iterations using only RGB images. Following \cite{langsplat,langsplatv2}, we extract language feature maps by generating segmentation masks with the SAM ViT-H model~\cite{sam} at three hierarchical semantic levels, and encoding each mask with OpenCLIP ViT-B/16~\cite{openclip} to obtain \(512\)-dimensional CLIP features. We train our 2D sparse coding algorithm at each semantic level for 4,000 epochs using the Adam optimizer~\cite{adam} with a learning rate of \(0.0007\), with codebook size \(L=64\) and \(K=4\) non-zero coefficients. The codebook is initialized with the k-means algorithm on CLIP-encoded masked regions. Once complete, we lift the sparse coefficient maps to 3D Gaussians using our sparse uplifting algorithm. For evaluation, we follow LangSplatV2~\cite{langsplatv2} to calculate similarity scores and generate relevancy masks between rendered language features and text queries from the CLIP text encoder. All experiments are conducted on an NVIDIA RTX A6000 GPU.

\begin{table*}[t]
\centering
\caption{Quantitative comparisons of open-vocabulary 3D object localization and 3D semantic segmentation on the LERF dataset.}
\label{tab:lerf_results}
\resizebox{\textwidth}{!}{
\begin{tabular}{lccccc  ccccc}
\toprule
\multirow[c]{2}{*}{\textbf{Method}} &
\multicolumn{5}{c}{\textbf{3D Object Localization} (mAcc $\%$)} &
\multicolumn{5}{c}{\textbf{3D Semantic Segmentation} (mIoU $\%$)} \\
\cmidrule(lr){2-6} \cmidrule(lr){7-11}
& Ramen & Teatime & Kitchen & Figurines & Overall
& Ramen & Teatime & Kitchen & Figurines & Overall \\
\midrule
GS-Grouping~\cite{gaussian_grouping}  & 32.4 & 69.5 & 50.0 & 44.6 & 49.1 & 26.4 & 54.0 & 31.3 & 34.6 & 36.6 \\
LEGaussian~\cite{LEGaussians}   & 69.0 & 79.7 & 63.6 & 57.1 & 67.4 & 20.2 & 32.3 & 22.3 & 23.4 & 24.6 \\
GOI~\cite{goi}          & 56.3 & 67.8 & 68.2 & 44.6 & 59.2 & 33.7 & 55.8 & 54.5 & 23.9 & 42.0 \\
GAGS~\cite{gags}         & 69.0 & 88.1 & \underline{90.9} & 78.6 & 81.7 & 46.8 & 60.3 & 55.8 & 53.6 & 54.1 \\
LangSplat~\cite{langsplat}    & 73.2 & 88.1 & \textbf{95.5} & 80.4 & \underline{84.3} & 51.2 & 65.1 & 44.5 & 44.7 & 51.4 \\
LangSplatV2~\cite{langsplatv2}   & \underline{74.7} & \textbf{93.2} & 86.4 & \underline{82.1} & 84.1 & \underline{51.8} & \underline{72.2} & 59.1 & 56.4 & 59.9 \\
Occam's LGS~\cite{occams} & \underline{74.7} & \textbf{93.2} & 81.8 & 80.4 & 82.5 & 51.0 & 70.2 & \textbf{65.3} & 58.6 & 61.3 \\
VALA~\cite{vala} & \textbf{75.6} & \underline{91.5} & 86.4 & \underline{82.1} & 83.9 & 51.5 & 70.2 & \underline{65.1} & \underline{59.9} & 
\underline{61.7} \\
\midrule
\textbf{Ours} & \underline{74.7} & \underline{91.5} & \textbf{95.5} & \textbf{85.7} & \textbf{86.9} & \textbf{57.6} & \textbf{75.3} & 64.4 & \textbf{60.4} & \textbf{64.4} \\

\bottomrule

\end{tabular}}

\end{table*}

\subsection{Quantitative Results on LERF-OVS}

\paragraph{Runtime and Memory Analysis.}

To evaluate the efficacy of the SCOUP framework, we recorded 3D semantic scene reconstruction runtimes, peak GPU memory allocation, and language feature rendering speeds across all three semantic levels. To ensure a rigorous evaluation, all baselines were re-measured locally on our GPU using synchronized CUDA clocks. As shown in Fig.~\ref{fig:pipeline}(d), under these standardized conditions, our method achieves a rendering speed of 29 FPS for all three semantic levels in parallel, matching the throughput of LangSplatV2~\cite{langsplatv2}. Crucially, our approach is more than \(400\times\) faster in reconstruction and \(3\times\) more memory efficient than LangSplatV2. While our reconstruction runtime is comparable to training-free methods like Occam's LGS~\cite{occams} and VALA~\cite{vala}, these alternatives store full CLIP embeddings into 3DGS. VALA additionally filters out roughly 90\% of Gaussians, yet still renders \(3\times\) slower than SCOUP. In contrast, SCOUP stores only four coefficients and their indices per level on all Gaussians from the underlying 3DGS, achieving an optimal balance between reconstruction speed, memory overhead, and rendering performance.

\paragraph{Main Results.} Table \ref{tab:lerf_results} presents the 3D open-vocabulary localization accuracy and segmentation IoU scores on the LERF dataset~\cite{LERF2023}. For the localization task, our method achieves superior performance on the "Figurines" scene while remaining highly competitive with state-of-the-art results across all other scenes, resulting in the highest overall localization accuracy. Regarding 3D open-vocabulary segmentation, our approach leads to substantial overall improvements, particularly on the "Ramen" and "Teatime" scenes, where it outperforms existing methods by a significant margin. Specifically, in a direct comparison with LangSplatV2~\cite{langsplatv2}, our method achieves a \(+4.5\%\) gain in IoU score. This improvement is realized while maintaining the same high-speed rendering capabilities and requiring significantly less time for semantic scene reconstruction. 

\begin{table*}[t]
\centering
\caption{Quantitative 3D semantic segmentation results on Mip-NeRF360 and 3D-OVS datasets.}
\label{tab:combined_semseg}
\resizebox{\textwidth}{!}{
\begin{tabular}{lccccccccccc}
\toprule
& \multicolumn{5}{c}{\textbf{Mip-NeRF360} (mIoU $\%$)} & \multicolumn{6}{c}{\textbf{3D-OVS} (mIoU $\%$)} \\
\cmidrule(lr){2-6} \cmidrule(lr){7-12}
\textbf{Method} 
& Room & Counter & Garden & Bonsai & Overall
& Bed & Bench & Room & Sofa & Lawn & Overall \\
\midrule
LEGaussian~\cite{LEGaussians} 
& 25.5 & 35.3 & 33.2 & 22.3 & 29.1
& 84.9 & 91.1 & 86.0 & 87.8 & 92.5 & 88.5 \\

GOI~\cite{goi} 
& 60.3 & 46.6 & \underline{59.8} & 67.3 & 58.5
& 89.4 & 92.8 & 91.3 & 85.6 & 94.1 & 90.6 \\

LangSplat~\cite{langsplat} 
& 53.2 & 68.8 & 51.9 & 55.4 & 57.3
& 92.5 & 94.2 & 94.1 & \underline{90.0} & 96.1 & 93.4 \\

LangSplatV2~\cite{langsplatv2} 
& 64.3 & \underline{75.1} & \textbf{65.0} & \textbf{73.1} & \underline{69.4}
& \underline{93.0} & \underline{94.9} & \underline{96.1} & \textbf{92.3} & 96.6 & \underline{94.6} \\

Occam's LGS~\cite{occams} 
& - & - & - & - & -
& \textbf{96.8} & \textbf{95.8} & \textbf{96.5} & 88.8 & \underline{97.0} & \textbf{95.0} \\

\midrule
\textbf{Ours} 
& \textbf{66.5} & \textbf{82.3} & 57.4 & \underline{71.7} & \textbf{69.5}
& 87.7 & \underline{94.9} & 95.1 & 88.2 & \textbf{97.2} & 92.6 \\
\bottomrule
\end{tabular}
}
\end{table*}

\begin{figure}[t]
  \centering
  \includegraphics[width=\linewidth]{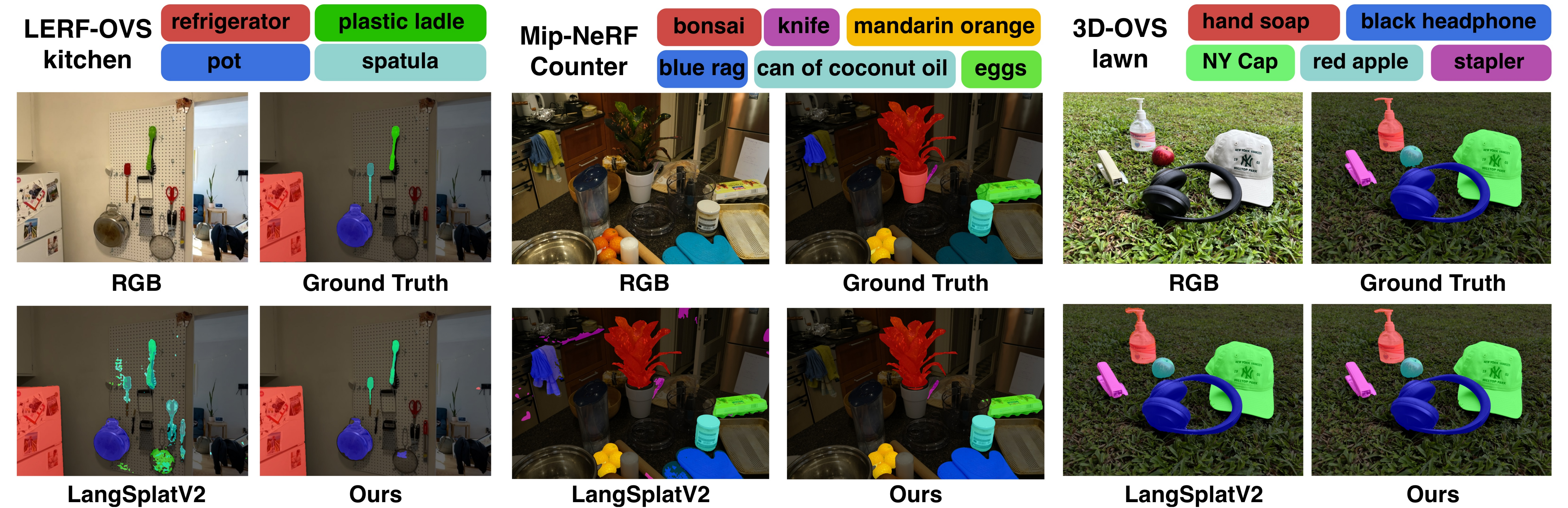}
  \caption{Qualitative results for open-vocabulary 3D object localization on LERF-OVS, Mip-NeRF360, and 3D-OVS datasets.}
  \label{fig:qualitative_seg}
\end{figure}

\begin{table}[t]
\centering
\caption{Semantic reconstruction runtime (2D$\rightarrow$3D), peak allocated GPU memory during reconstruction, and language rendering speed (3D$\rightarrow$2D) for all three levels on 3D-OVS and Mip-NeRF360 datasets. Experiments were run on a NVIDIA RTX A6000.}
\label{tab:runtime_memory_3dovs_mipnerf}
\resizebox{\textwidth}{!}{
\begin{tabular}{l cc  cc  cc }
\toprule
\multirow{2}{*}{\textbf{Method}} &
\multicolumn{2}{c}{\textbf{2D$\rightarrow$3D} (min)} &
\multicolumn{2}{c}{\textbf{GPU Mem.} (GB)} &
\multicolumn{2}{c}{\textbf{3D$\rightarrow$2D} (FPS)} \\
\cmidrule(lr){2-3} \cmidrule(lr){4-5} \cmidrule(lr){6-7}
& Mip-NeRF360 & 3D-OVS
& Mip-NeRF360 & 3D-OVS
& Mip-NeRF360 & 3D-OVS \\
\midrule
LangSplatV2~\cite{langsplatv2}   & 684 & 570 & 37.7 & 28.5  & \textbf{15}  & \textbf{15}   \\
\textbf{Ours}                    & \textbf{1.4} & \textbf{0.5} & \textbf{10.4} & \textbf{4.4} & \textbf{15} & \textbf{15}  \\
\bottomrule
\end{tabular}}
\end{table}

\begin{table}[ht]
\centering
\caption{Ablation study evaluating the impact of dense feature uplifting, sparse code uplifting, and Top-$K$ filtering on LERF-OVS dataset.}
{\setlength{\tabcolsep}{6pt} 
\renewcommand{\arraystretch}{1.1} 
\resizebox{\linewidth}{!}{
\begin{tabular}{ccc|ccccc}
\toprule
Feature uplifting & Sparse uplifting & +Top-K & mIoU (\%) & mAcc (\%) &  2D$\rightarrow$3D (min) & 3D$\rightarrow$2D (FPS) & Storage (GB)\\
\midrule
\checkmark &  &  & 60.5 & 82.0 & 21.3 & 0.5 & 7.8 \\
- & \checkmark & & 60.0 & 81.2 & 0.7 & 22.0 & 1.0 \\
- & \checkmark & \checkmark & \textbf{64.4} & \textbf{86.9} & \textbf{0.7} & \textbf{29.3} & \textbf{0.1} \\
\bottomrule
\end{tabular}}}
\label{tab:main_ablation}
\end{table}

\subsection{Quantitative Results on 3D-OVS and Mip-NeRF360}

\paragraph{Main results.} Table \ref{tab:combined_semseg} shows 3D open-vocabulary segmentation results on the Mip-NeRF360 and 3D-OVS datasets. Our method achieves performance comparable to state-of-the-art baselines across the majority of scenes. The notable exception is the 3D-OVS: ``Bed'' scene, where performance is affected by a single query: ``black leather shoe.'' On this scene, SAM~\cite{sam} occasionally fails to disentangle the ``shoe'' from the surrounding ``white sheet'' region in the preprocessing step of 3DLGS. Since the 3D-OVS evaluation set contains only a small number of test queries per scene, this single failure substantially lowers the reported mIoU. Excluding this query increases our method's mean IoU for that scene to \(96.4\%\). 

\paragraph{Runtime and memory analysis.} 

As shown in Table~\ref{tab:runtime_memory_3dovs_mipnerf}, SCOUP significantly outperforms LangSplatV2~\cite{langsplatv2} in both reconstruction runtime and GPU memory footprint while matching its state-of-the-art rendering speed. This unique combination of rapid 3D semantic reconstruction, low memory overhead, and high rendering performance makes our approach a highly practical system for real-world applications.

\subsection{Ablation Study}

In Table \ref{tab:main_ablation}, we evaluate the impact of our Sparse Code Uplifting and Top-\(K\) Filtering components on the LERF-OVS dataset. Dense feature uplifting takes approximately 21 minutes on average across four scenes to uplift all three semantic levels. Our sparse uplifting drastically reduces this process to under 1 minute while maintaining performance comparable to full CLIP feature aggregation. 

Furthermore, applying Top-\(K\) filtering to the coefficients of each Gaussian after uplifting results in absolute performance gains of +3.9\% and +4.9\% in IoU score and mean accuracy, respectively, compared to using full CLIP embeddings. This demonstrates that Top-\(K\) filtering effectively acts as a denoising mechanism that improves multi-view consistency. We include further ablation studies in Appendix~\ref{app:abl} and additional details on runtime analysis in Appendix~\ref{app:detailed_reconstruction_runtime}.

\subsection{Qualitative results} 

In Fig. \ref{fig:qualitative_loc}, we present qualitative results for open-vocabulary 3D object localization on the LERF-OVS dataset. While all two methods demonstrate strong performance in localizing target objects, SCOUP results in noticeably higher-confidence heatmaps and clearer object boundaries compared to LangSplatV2~\cite{langsplatv2}. Furthermore, Fig. \ref{fig:qualitative_seg} visualizes open-vocabulary 3D object segmentation on LERF-OVS Mip-NeRF360, and 3D-OVS.
In contrast to the 3D codebook optimization in LangSplatV2~\cite{langsplatv2}, in our 2D sparse coding, small and large objects contribute with more balance to the codebook learning process. This is particularly evident for challenging queries such as the ``spatula'' in the LERF-OVS kitchen scene and the ``knife'' in the Mip-NeRF360 counter scene. We include more qualitative results in Appendix \ref{app:qual}.

\section{Conclusion}
\label{sec:conclusion}

We present SCOUP, a framework for 3D Language Gaussian Splatting that achieves fast training, efficient storage, and fast rendering simultaneously by shifting sparse-code and codebook learning from 3D Gaussian space to 2D image space, then uplifting the learned sparse codes to the underlying 3D Gaussians through weighted sparse aggregation followed by top-\(K\) filtering, which suppresses multi-view inconsistencies. Compared to LangSplatV2, SCOUP achieves up to \(400\times\) training speedup and \(3\times\) lower training memory while maintaining state-of-the-art rendering speed and preserving storage efficiency and accuracy across multiple benchmarks. By reducing per-scene semantic reconstruction from hours to under a minute, SCOUP lowers the practical barrier to deploying open-vocabulary 3D scene understanding in real-world applications.

\bibliography{main}
\bibliographystyle{plainnat}


\appendix
\section{Algorithm}

The proposed sparse code uplifting procedure is summarized in Algorithm~\ref{alg:scu}.

\begin{algorithm}[t]
\caption{Sparse Code Uplifting}
\label{alg:scu}
\begin{algorithmic}
    \For{Gaussian point $i \in \mathcal{N}$}
        \State Initialize $\mathbf{w}_i \in \mathbb{R}^{L}$ as the zero vector
    \EndFor

    \For{frame $v \in \mathcal{V}$}
        \For{pixel $p \in \Omega_v$}
            \State $(\text{indices }\{j_1,\ldots,j_K\}, \text{ coefficients }\{c_{j_1},\ldots,c_{j_K}\})  \gets \operatorname{TopK}(\mathbf{W}_v(p))$
            \For{Gaussian point $i \in \mathcal{N}_{v,p}$}
                \For{each $j \in \{j_1,\dots,j_K\}$}
                    \State $\mathbf{w}_i[j] \gets \mathbf{w}_i[j] + c_j \cdot e_i(v,p)$
                \EndFor
            \EndFor
        \EndFor
    \EndFor

    \For{Gaussian point $i \in \mathcal{N}$}
        \If{$\mathbf{w}_i \neq \mathbf{0}$}
            \State $\hat{\mathbf{w}}_i = \text{Top}_K(\mathbf{w}_i) / \|\text{Top}_K(\mathbf{w}_i)\|_1$
        \EndIf
    \EndFor
\end{algorithmic}
\end{algorithm}

\section{Additional Evaluation Details}

We follow \cite{langsplat,langsplatv2} to compute relevancy scores between rendered language features and text queries. Specifically, the relevancy score between the rendered language feature at pixel position $p$, denoted by $\mathbf{F}_v(p)$, and the CLIP text embedding of the query, $\mathbf{t}_{\mathrm{query}}$, is defined as
\begin{equation}
R(p) = \min_k \frac{\exp\!\left(\alpha \cdot \mathbf{F}_v(p) \cdot \mathbf{t}_{\mathrm{query}}\right)}
{\exp\!\left(\alpha \cdot \mathbf{F}_v(p) \cdot \mathbf{t}_{\mathrm{query}}\right) + \exp\!\left(\alpha \cdot \mathbf{F}_v(p) \cdot \mathbf{t}^{k}_{\mathrm{canon}}\right)},
\end{equation}
where $\alpha$ is a temperature parameter set to 10, and $\mathbf{t}^{k}_{\mathrm{canon}}$ is the CLIP embedding of a predefined canonical phrase selected from ``object,'' ``things,'' ``stuff,'' and ``texture.'' 

For relevancy map post-processing, we follow LangSplatV2~\cite{langsplatv2}, modifying only the mask thresholds and the average-pooling kernel size. We use thresholds of 0.6, 0.45, and 0.3 on LERF~\cite{LERF2023}, Mip-NeRF360~\cite{mipnerf360}, and 3D-OVS~\cite{3dovs2023}, respectively, and set the average-pooling kernel size to 7 for all three datasets.

\section{Additional Ablation Studies}
\label{app:abl}

\subsection{Sparse Coding Parameters}

\begin{table}[t]
\centering
\small

\begin{minipage}[t]{0.45\linewidth}
\centering
\caption{Comparison for varying $L$. We report mean accuracy (\%) and mean IoU (\%) on LERF dataset.}
\label{tab:l_comparison}
\resizebox{0.8\linewidth}{!}{
\begin{tabular}{lccc}
\toprule
 $L$ & $32$ & $64$ & $128$  \\
\midrule
mean Accuracy (\%) & 77.2 & \textbf{86.9} & \underline{85.1} \\
mean IoU (\%) & 60.6 & \underline{64.4} & \textbf{65.5} \\
\bottomrule
\end{tabular}}
\end{minipage}
\hfill
\begin{minipage}[t]{0.45\linewidth}
\centering
\caption{Comparison for varying $K$. We report mean accuracy (\%) and mean IoU (\%) on LERF dataset.}
\label{tab:k_comparison}
\resizebox{0.8\linewidth}{!}{
\begin{tabular}{lccc}
\toprule
 $K$ & $2$ & $4$ & $8$  \\
\midrule
mean Accuracy (\%) & \underline{82.9} & \textbf{86.9} & 78.1 \\
mean IoU (\%) & 62.5 & \textbf{64.4} & \underline{63.8} \\
\bottomrule
\end{tabular}}
\end{minipage}

\end{table}

In Table \ref{tab:l_comparison} and Table \ref{tab:k_comparison}, we ablate the codebook size ($L$) and sparsity level ($K$) on the LERF-OVS dataset, respectively. We find that $L=64$ provides the best balance between open-vocabulary performance and rendering efficiency, as larger codebooks introduce additional decoding overhead. Furthermore, setting $K=4$ during both 2D sparse coding and Top-$K$ filtering yields the highest mean accuracy, indicating that a larger $K$ introduces semantic ambiguity. While these optimal parameters are consistent with those reported in LangSplatV2~\cite{langsplatv2}, our framework learns the codebook entirely in the 2D domain. 

\begin{table}[t]
\centering
\caption{Effect of entropy-based Gaussian filtering on the LERF-OVS dataset.}
\label{tab:entropy_ablation}
\begin{tabular}{lcc}
\toprule
Filtering strategy & mIoU (\%) & mAcc (\%) \\
\midrule
No filtering & \textbf{64.4} & \textbf{86.9} \\
Remove 50\% of the highest-entropy Gaussians by $H(\mathbf{w}_i)$ & \underline{61.0} & \underline{86.5} \\
Remove 50\% of the lowest-entropy Gaussians by $H(\mathbf{w}_i)$ & 33.1 & 78.2 \\
\bottomrule
\end{tabular}
\end{table}

\subsection{Entropy-based Analysis of Top-K Filtering}

In Table \ref{tab:entropy_ablation}, we study whether entropy can be used to assess the importance of Gaussians before Top-$K$ filtering. For each Gaussian, we compute the entropy of its coefficient distribution as
\begin{equation}
H(\mathbf{w}_i) = - \sum_{l=1}^{L} w_{i,l} \log w_{i,l}.
\end{equation}
We then filter out 50\% of the Gaussians based on their entropy values to evaluate their contribution to open-vocabulary 3D object localization and segmentation.

The results indicate that Gaussians with low-entropy coefficients are considerably more important. Since low entropy reflects a more confident and concentrated distribution during multi-view aggregation, these Gaussians provide more reliable semantic cues. Removing 50\% of the low-entropy Gaussians leads to a substantial drop in mean IoU and mean accuracy. By contrast, removing 50\% of the high-entropy Gaussians results in only a minor performance decrease. These results support the consensus interpretation underlying our Top-K filtering: Gaussians whose aggregated coefficients are concentrated on a few atoms encode more reliable multi-view semantics than those with diffuse distributions.

\section{Detailed Runtime Analysis}
\label{app:detailed_reconstruction_runtime}

\begin{figure}[t]
    \centering
    \begin{subfigure}{0.54\linewidth}
        \centering
        \includegraphics[width=\linewidth]{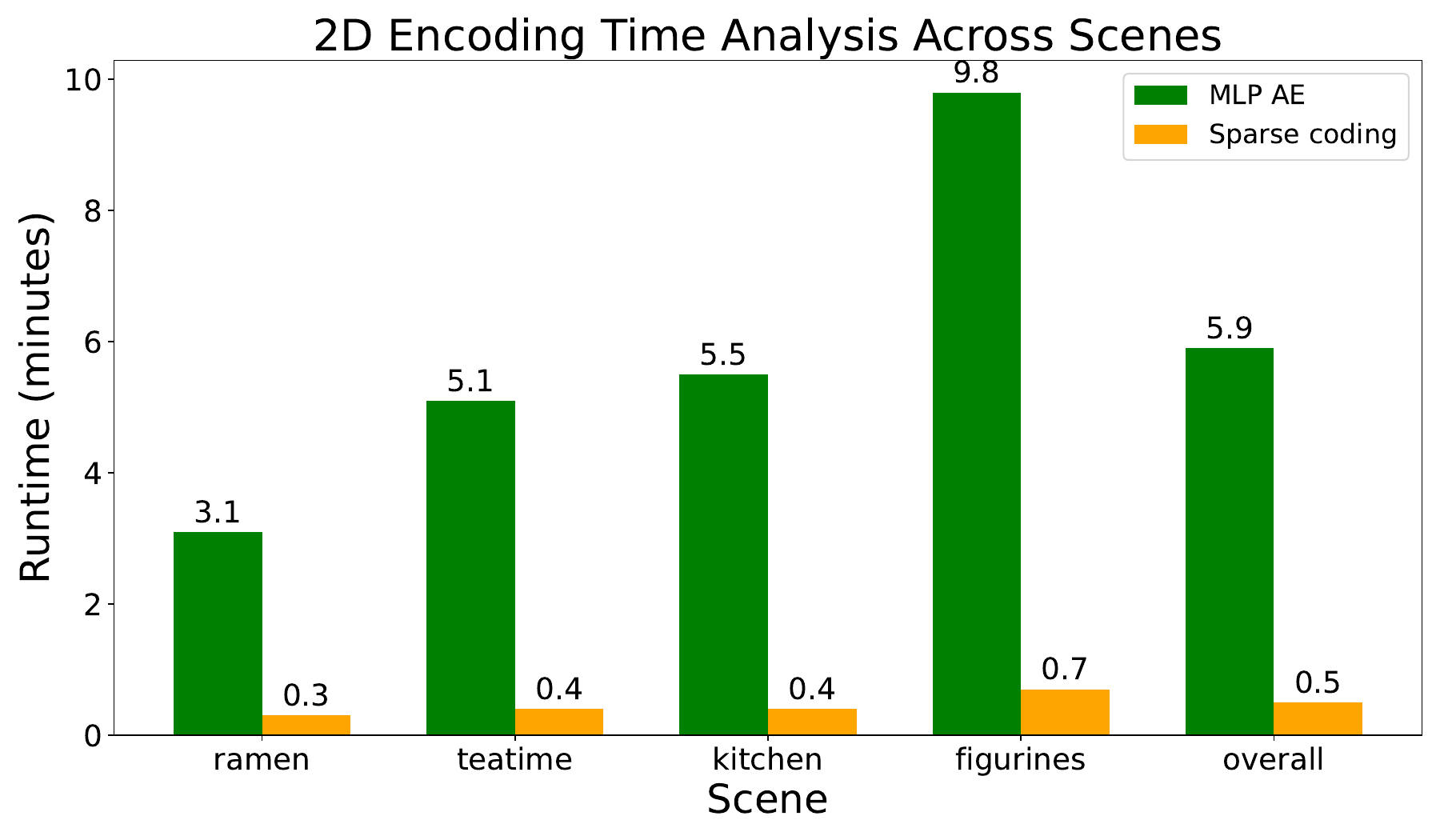}
        \label{fig:encoding-time}
    \end{subfigure}
    \hfill
    \begin{subfigure}{0.42\linewidth}
        \centering
        \includegraphics[width=\linewidth]{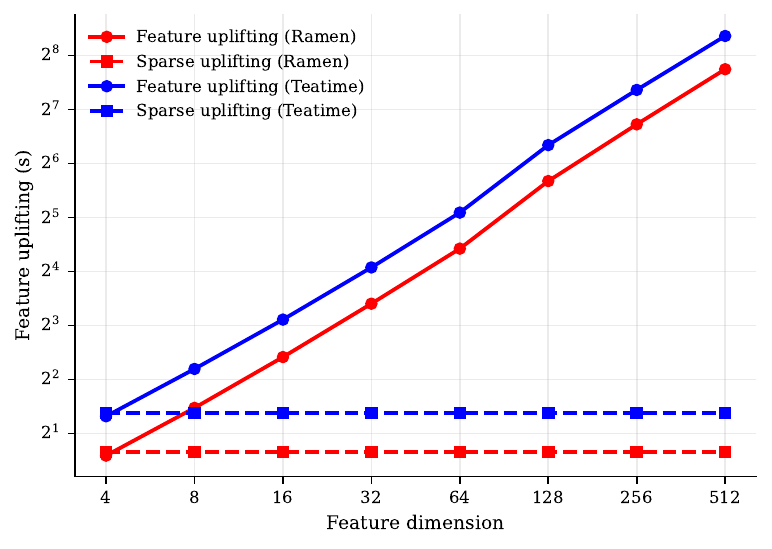}
        \label{fig:uplift_time}
    \end{subfigure}
    \caption{
    Left: 2D encoding runtime of the MLP autoencoder from LangSplat~\cite{langsplat} compared to our 2D sparse coding. Right: sparse coefficient uplifting runtime of SCOUP compared to full-dimensional feature uplifting on the LERF-OVS dataset (NVIDIA RTX A6000 GPU).
    }
    \label{fig:detailed_runtime_comparison}
\end{figure}

In Fig. \ref{fig:detailed_runtime_comparison} (left), we compare our 2D Sparse Coding algorithm against the MLP autoencoder used in LangSplat~\cite{langsplat}. Our approach is over $10\times$ faster at encoding CLIP language embeddings for each SAM-segmented region while avoiding the information loss typically associated with MLP-based compression.

Furthermore, we analyze the scalability of our sparse code uplifting on the LERF-OVS dataset. As shown in Fig. \ref{fig:detailed_runtime_comparison} (right), uplifting sparse coefficient maps requires only 2.5 seconds and 1.5 seconds per semantic level on the Teatime and Ramen scenes, respectively, regardless of the feature dimensionality. This indicates that approximately 80\% of the total computational runtime in 3D semantic reconstruction stems from the 2D sparse coding of the feature maps, a step we can efficiently compute prior to the uplifting process.

\section{Multi-View Consistency from 2D Sparse Coding}

\begin{figure}[t]
    \centering
    \includegraphics[width=0.5\linewidth]{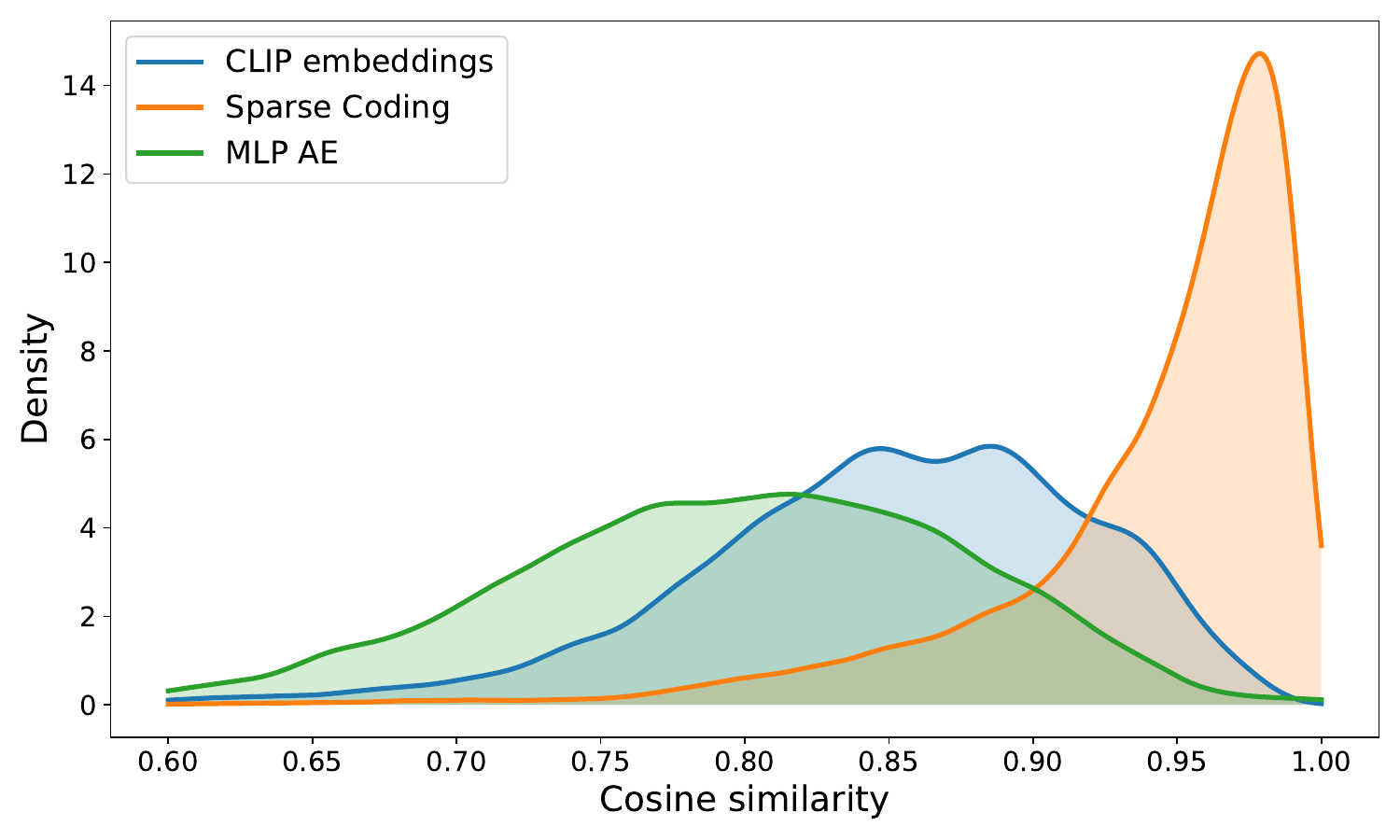}
    \caption{Comparison of multi-view consistency among ground-truth CLIP features, decoded MLP autoencoder features from LangSplat~\cite{langsplat}, and decoded sparse coding features from SCOUP on the \textit{Teatime} scene from the LERF~\cite{LERF2023} dataset.}
    \label{fig:multi-view-cons}
\end{figure}

In Fig. \ref{fig:multi-view-cons}, we evaluate the multi-view consistency of three types of features on the coarse semantic level of the \textit{Teatime} scene from LERF-OVS~\cite{LERF2023}: ground-truth CLIP features, decoded features from the 2D autoencoder of LangSplat~\cite{langsplat}, and decoded features from the 2D sparse coding method used in SCOUP. To measure multi-view consistency, we use sparse 3D points from COLMAP~\cite{schoenberger2016sfm,schoenberger2016mvs} that we used to initialize 3DGS training. For each sparse 3D point, we identify its corresponding 2D observations in multiple views and compute the cosine similarity between the associated 2D features. The results indicate that 2D sparse coding improves feature consistency across views by increasing the cosine similarity between matched 2D features. By contrast, the MLP autoencoder may lead to a loss of semantic information due to the aggressive compression of CLIP embeddings, resulting in smaller cosine similarity between corresponding 2D features. These results justify our adoption of 2D sparse coding, as it preserves semantic structure across views more faithfully than the autoencoder baseline, while also improving training efficiency, as shown in Appendix~\ref{app:detailed_reconstruction_runtime}.

\section{Limitations and Broader Impact}

\begin{figure}[t]
  \centering
  \includegraphics[width=0.9\linewidth]{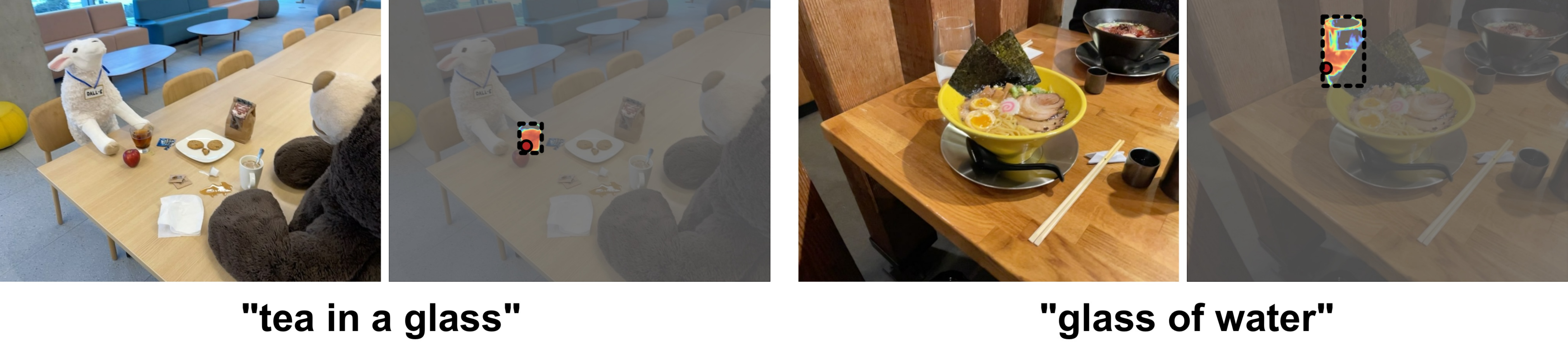}
  \caption{Limitation of direct feature uplifting in 3DLGS.}
  \label{fig:glass_study}
\end{figure}

Despite its efficiency, SCOUP inherits the computational overhead of the large visual foundation models used for pre-processing. Specifically, the heavy feature extraction required by SAM and CLIP remains a bottleneck in pre-processing. Furthermore, since our sparse feature uplifting relies on 3DGS priors optimized for visual appearance, objects with high transparency pose a challenge. Gaussians representing transparent surfaces often have lower opacity values, leading to less confident semantic heatmaps on objects like glass, as shown in Fig.~\ref{fig:glass_study}.

Along with the computational overhead of VLMs, our method inherits their semantic biases, which propagate to the downstream task. At the same time, SCOUP's efficiency carries a positive environmental impact: it makes per-scene open-vocabulary 3D reconstruction substantially cheaper, reducing semantic reconstruction time from hours to under a minute and lowering memory and GPU utilization during training.

\section{Additional Qualitative Results}
\label{app:qual}

\begin{figure}[t]
  \centering
  \includegraphics[width=0.9\linewidth]{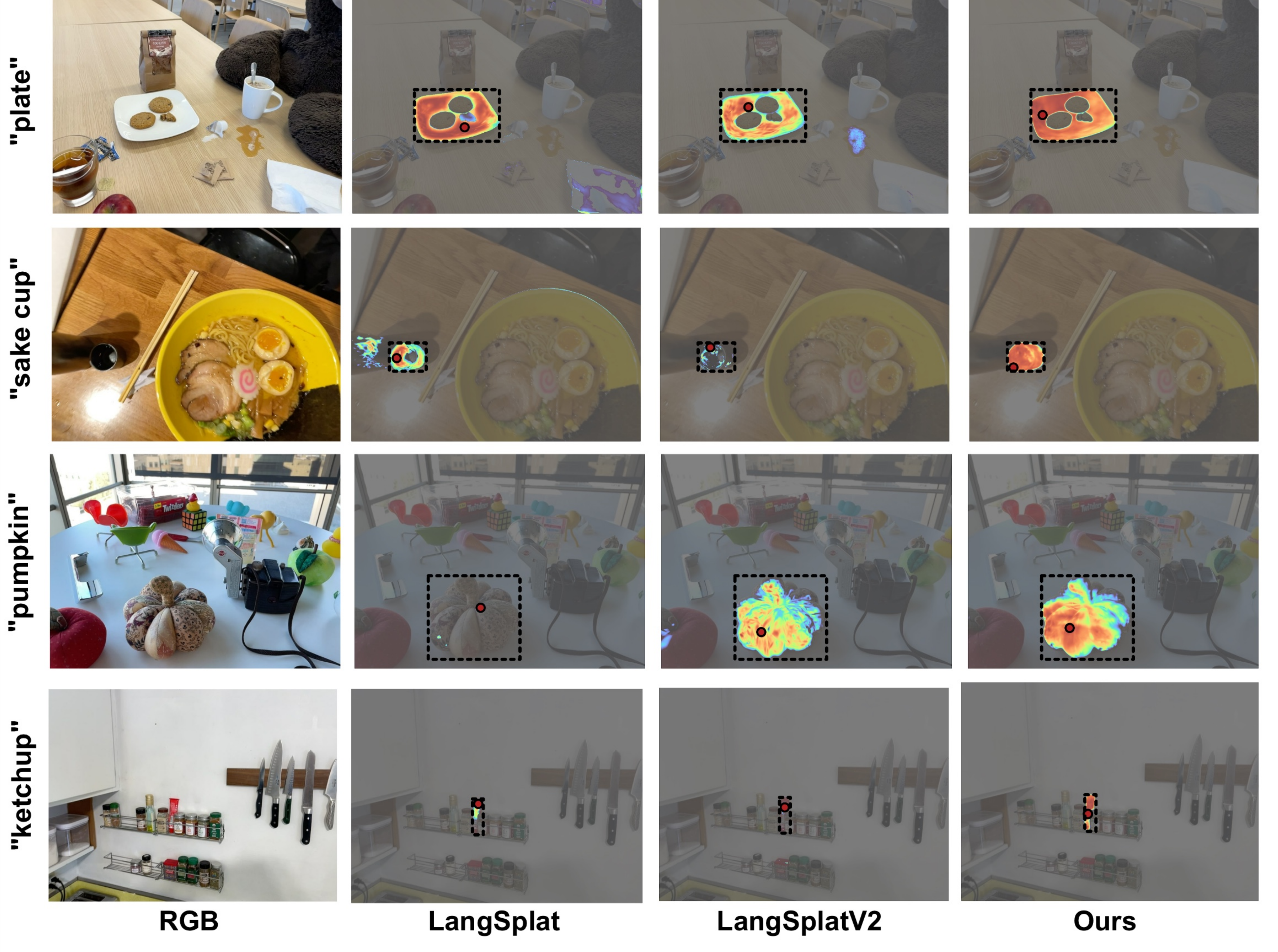}
  \caption{Additional qualitative results for open-vocabulary 3D object localization on the LERF-OVS dataset.}
  \label{fig:qualitative_loc_supp}
\end{figure}

We present additional qualitative examples of open-vocabulary 3D semantic localization on the LERF-OVS~\cite{LERF2023} dataset in Fig. \ref{fig:qualitative_loc_supp}.

\begin{figure}[t]
  \centering
  \includegraphics[width=0.9\linewidth]{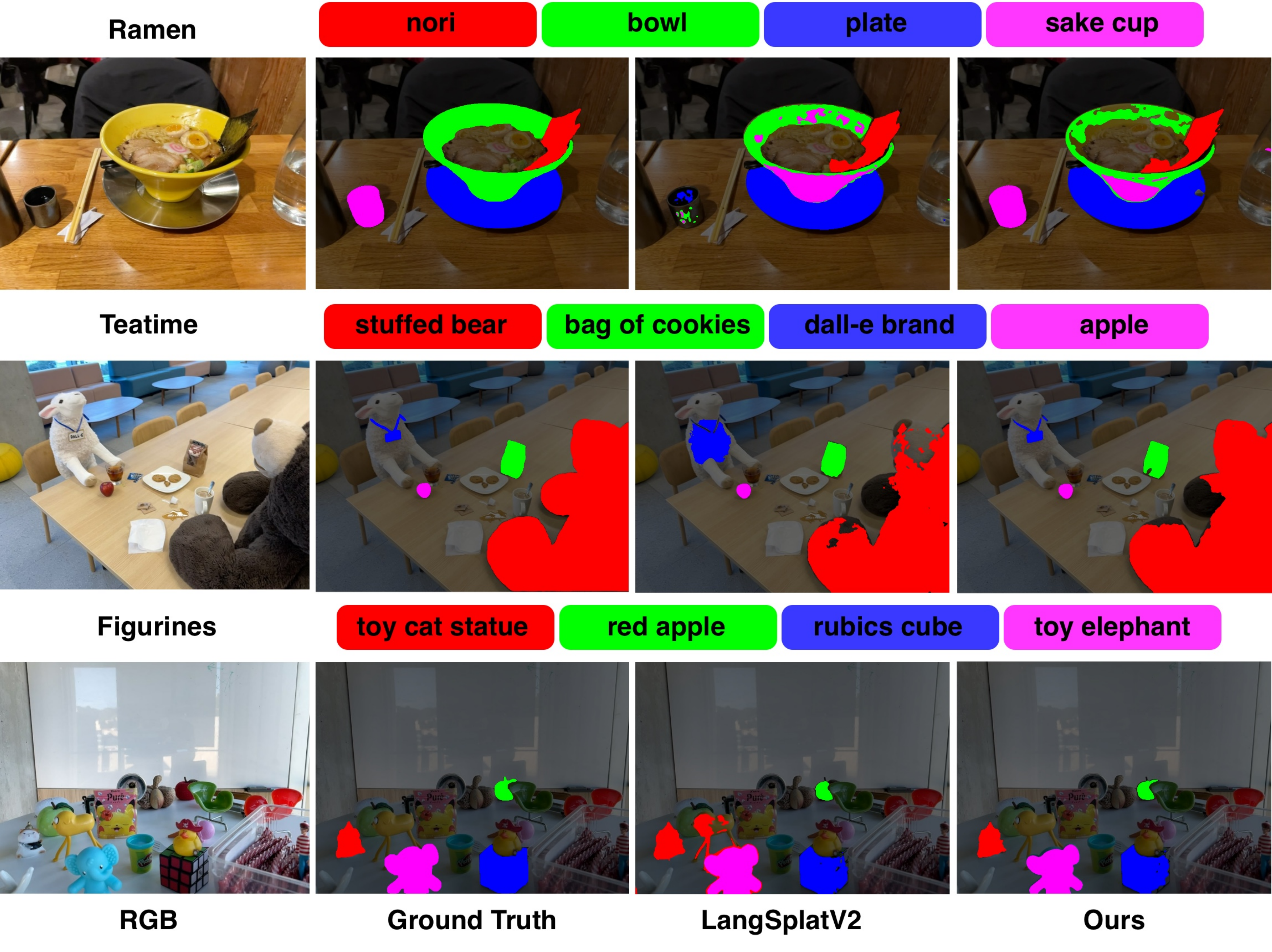}
  \caption{Additional qualitative results for open-vocabulary 3D object segmentation on the LERF-OVS dataset.}
  \label{fig:qualitative_seg_lerf}
\end{figure}

\begin{figure}[t]
  \centering
  \includegraphics[width=0.9\linewidth]{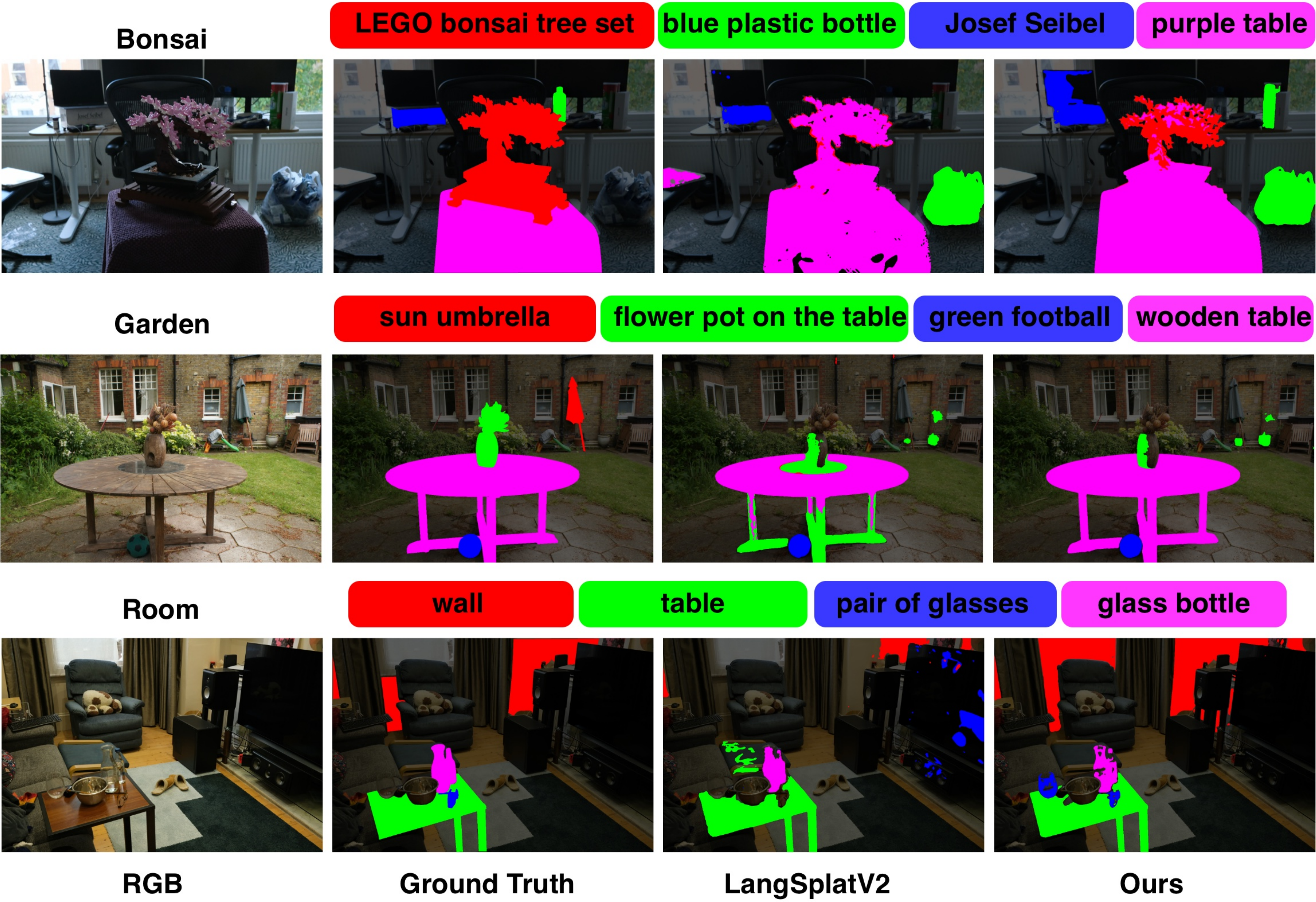}
  \caption{Additional qualitative results for open-vocabulary 3D object segmentation on the Mip-NeRF360 dataset.}
  \label{fig:qualitative_seg_mipnerf}
\end{figure}

\begin{figure}[t]
  \centering
  \includegraphics[width=0.9\linewidth]{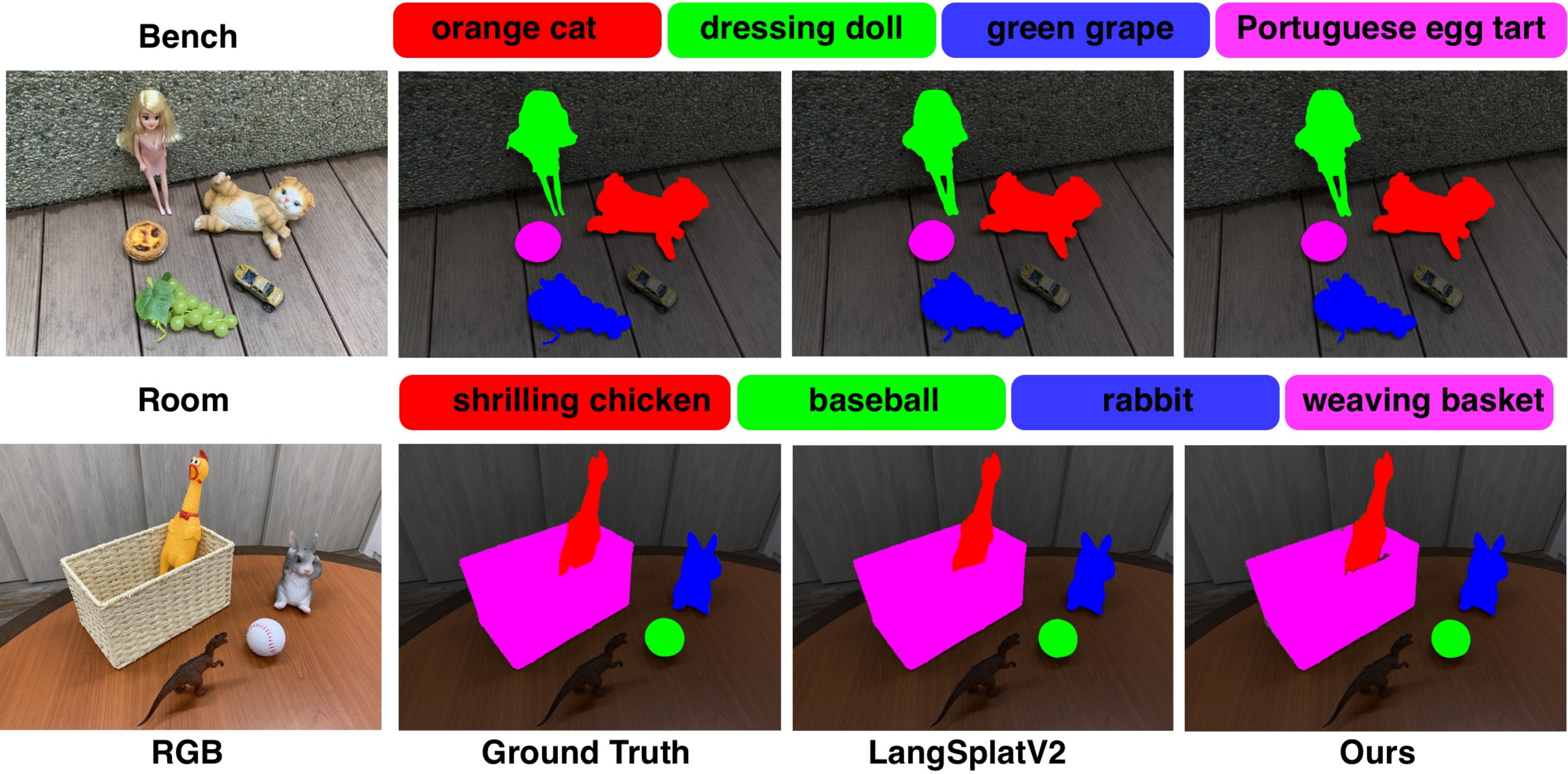}
  \caption{Additional qualitative results for open-vocabulary 3D object segmentation on the 3D-OVS dataset.}
  \label{fig:qualitative_seg_3dovs}
\end{figure}

We present additional qualitative examples of open-vocabulary 3D semantic segmentation on the LERF-OVS~\cite{LERF2023}, Mip-NeRF360~\cite{mipnerf360}, and 3D-OVS~\cite{3dovs2023} datasets in Fig. \ref{fig:qualitative_seg_lerf}, Fig. \ref{fig:qualitative_seg_mipnerf}, and Fig. \ref{fig:qualitative_seg_3dovs}, respectively.

\end{document}